\documentclass{article}
\usepackage[ruled,linesnumbered,boxed]{algorithm2e}
\usepackage{bm}
\usepackage{times,authblk}
\usepackage{natbib}
\usepackage{mathtools}
\usepackage{booktabs} 
\newcommand{\maxrev}{\textsc{Max-Rev} }
\newcommand{\maxrevts}{\textsc{Max-Rev-TS} }
\newcommand{\maxrevps}{\textsc{Max-Rev-Passive} }
\newcommand{\dit}{d_{i,t}}
\newcommand{\cA}{\mathcal{A}}
\newcommand{\cB}{\mathcal{B}}
\newcommand{\dov}{\bm{f}}
\newcommand{\doit}{f_{i,t}}
\newcommand{\doio}{f_{i,1}}
\newcommand{\doBt}{f_{B,t}}
\newcommand{\doot}{f_{1,t}}
\newcommand{\pit}{p_{i, t}}
\newcommand{\poit}{p_{i, t-1}}
\newcommand{\Rt}{R_t}

\newcommand{\Revit}{Rev_{i,t}}
\newcommand{\Revt}{Rev_{t}}
\newcommand{\pt}{\mathbf{p}_t}
\newcommand{\p}{\mathbf{p}}
\newcommand{\cC}{\mathcal{C}}
\newcommand{\cCt}{\mathcal{C}_t}
\newcommand{\gamstar}{\mathbf{\gamma}_{\star}}
\newcommand{\Rtbar}{\bar{R}_t}
\newcommand{\cN}{N}
\newcommand{\vp}{\mathbf{p}}

\newcommand{\vtheta}{\bm{\theta}}
\newcommand{\vgamma}{\bm{\gamma}}
\newcommand{\gamstari}{\gamma_{\star,i}}
\newcommand{\vbeta}{\bm{\beta}}
\newcommand{\betat}{\mathbf{\beta}_{t}}

\newcommand{\mM}{\bm{M}}

\newcommand{\vx}{\bm{x}}
\newcommand{\mSigma}{\bm{\Sigma}}
\newcommand{\vmu}{\bm{\mu}}
\newcommand{\cG}{\mathcal{G}}
\newcommand{\bbE}{\bm{E}}
\newcommand{\bbR}{\bm{R}}

\newcommand{\Bo}{\cB_1}

\newcommand{\Bt}{\cB_2}
\newcommand{\Bp}{\cB_{p}}
\newcommand{\Bm}{\cB_{\maxrev}}
\newcommand{\Bf}{\cB_{f}}
\newcommand{\Bts}{\cB_{TS}}

\renewcommand{\pi}{p_i}

\newenvironment{game}[1][htb]
  {
   \begin{algorithm}[#1]%
  }{\end{algorithm}}

\title{Thompson Sampling for Dynamic Pricing}
\begin{document}


\author[]{Ravi Ganti\thanks{rmahapatruni@walmartlabs.com}}
\author[]{Matyas Sustik}
\author[]{Quoc Tran}
\author[]{Brian Seaman}
\affil[]{Walmart Labs, San Bruno, CA, USA}
\maketitle
\date{}
\begin{abstract}
In this paper we apply active learning algorithms for dynamic pricing in a prominent e-commerce website. Dynamic pricing involves changing the price of items on a regular basis, and uses the feedback from the pricing decisions to update prices of the items. Most popular approaches to dynamic pricing use a passive learning approach, where the algorithm uses historical data to learn various parameters of the pricing problem, and uses the updated parameters to generate a new set of prices. We show that one can use active learning algorithms such as Thompson sampling to more efficiently learn the underlying parameters in a pricing problem. We apply our algorithms to a real e-commerce system and show that the algorithms indeed improve revenue compared to pricing algorithms that use passive learning.
\end{abstract}

%
%



\section{Introduction}
The Internet has enabled many industries to go online. e-commerce which entails selling of goods both physical and digital, via the Internet, is one of the vibrant industries in modern times. The presence of gigantic all-purpose e-commerce retailers such as Walmart.com, Amazon.com, Jet.com, Target.com, and the more specialized and niche e-commerce retailers such as Moosejaw, Modcloth.com, Shoebuy.com, Wine.com, Bonobos.com has shifted commerce from offline medium to online medium. While the biggest brick-and-mortar stores are constrained by shelf space and tend to stock only the 100K most popular items, the largest of e-commerce retailers are less constrained by shelf space and tend to carry more than 400 million items. e-commerce retailers have also benefited from the frictionless, digital economy, and the explosion of online activity in general. It is now not hard to study customer behaviour online such as what items a customer viewed on a website, what items are popular in general, what items a customer bought in the past, and use this personalized information along with information from other sources to come up with actionable insights to entice and retain customers.

One of the most important problem any retailer has to solve is, how to price items correctly. Setting the prices too high or too low can cause bad customer experience and dramatically effect the bottom-line of the retailer. The size of the catalog carried by a brick-and-mortar store is very small, and because of the difficulty in changing prices, pricing of an items is an easy problem and can be solved by carrying out elementary data analysis in excel. However, the presence of a large collection of products on an e-commerce website, the ability to collect large amounts of user behaviour data easily, and also the relative ease with which price changes can be made on a large scale, makes the problem of pricing both challenging and rich. In traditional brick-and-mortar stores price of items tend to remain unchanged. However, such static pricing policies do not work in an e-commerce setting. Static pricing does not simulate the demand/price curve, nor does it allow a firm to respond to competitor price changes. Moreover static pricing is unable to fully exploit the power of computation, data, and the ease of implementing price changes in an online world, As a result, dynamic pricing is the pricing policy of choice in an e-commerce setting.

In this paper we study algorithms for dynamic pricing. Dynamic pricing in an e-commerce setting is a hard problem because of the scale of the problem and uncertain user behaviour. It is not clear how one can use the past collected data to come up with better prices. A typical approach to dynamic pricing consists of performing the following four steps: (i) Collect historical data about the price of an item and the demand of item at different price points, (ii) Posit a statistical model for the demand as a function of price, and estimate the model parameters using historical data, (iii) Using the learned demand function, optimize some metric of interest (e.g. revenue) to get the new optimal price, (iv) apply the obtained optimal price to the item for the next few days and repeat the above. We classify these approaches to dynamic pricing as passive learning approaches. Such approaches are myopic and try to optimize for the metric in the short term without trying to make an active effort to learn the demand function. In fact such approaches have shown to lead to incomplete learning~\citep{harrison2012bayesian}, and have poor performance, and lose revenue on the long run. In this paper we show how one can avoid this problem. Our contributions are as follows.
\begin{enumerate}
  \item \textit{We propose and implement a simple, scalable, active learning algorithm for dynamic pricing.} While passive learning based dynamic pricing approaches treat the pricing problem as a vanilla optimization problem of optimizing the cumulative metric of interest over a finite horizon $T$, we introduce an algorithm called \maxrevts that views dynamic pricing problem as an optimization problem under uncertainty.
  \item We begin by explaining the architecture of a dynamic pricing system that is currently in place. Our DP system makes precise parametric assumptions on the demand function and maximizes a metric of interest daily under constraints. To make our discourse concrete we shall use the revenue function as the metric of interest in this paper~\footnote{We shall use the words revenue and reward interchangeably.}. The parameters of the demand function are updated regularly. Since this DP system is based on passive learning, we shall call this system as \maxrevps.
  \item We then introduce a new DP system, \maxrevts which is very similar to \maxrevps, but one which uses a multi-armed bandit (MAB) algorithm called Thompson sampling~\citep{thompson1933likelihood}. \maxrevts uses the same parametric formula for the demand function as \maxrevps and proceeds to maximize revenue over a fixed horizon. TS operates as follows: A Bayesian prior is put on the unknown parameters of the demand function model and using the Bayes rule the posterior is derived. We use a conjugate prior-likelihood pair to obtain posterior updates in closed form. TS then samples a parameter from the posterior distribution and derive a new set of prices by maximizing the reward function for the next time step using the sampled parameters. This process is repeated until the end of the horizon.
  \item The main focus of this paper is provide an implementation of \maxrevts in a major e-commerce system. We discuss various practical issues that we had to take care of when applying TS in a real production environment. We show results of our implementation in a $5$ week period and compare it to the \maxrevps algorithm. The results show that \maxrevts shows a statistically significant improvement over passive learning algorithms for dynamic pricing under appropriate conditions. To the best of our knowledge we are the first to demonstrate active DP algorithms in a real e-commerce production system. Furthermore, while typical applications of MAB algorithms in production systems have dealt with the case when there are only finite number of arms~\cite{agarwal2016multiworld, shah2017practical}, our paper is the first to demonstrate practical application of a more involved bandit optimization problem which has infinite arms.
  \end{enumerate}


\section{Related Work}
DP has been an active area of research for more than two decades and has seen considerable amount of attention in operations research (OR), management sciences (MS), marketing, economics and computer sciences. A lot of work in dynamic pricing was traditionally carried out in OR/MS, and economics communities. It was pioneered by airline industry, where prices are adjusted over time via capacity control, and dynamic programming solutions are used to decide the fares~\cite{talluri2006theory,phillips2005pricing}.  DP has been considered under various settings such as pricing with and without inventory constraints, pricing with single or multiple items, finite or infinite horizon DP, DP with known or unknown demand function. It is not possible to give a thorough literature survey of dynamic pricing. We refer the interested reader to the excellent survey of ~\cite{den2015dynamic} for a comprehensive survey on DP. In this section we give a limited survey of DP.

In economics, the early works of~\cite{rothschild1974two, aghion1991optimal} considered the problem of DP with unknown demand model. In these papers, the dynamic pricing problem is cast as a decision making problem in an infinite horizon, with discounted rewards. A Bayesian setting is used and a solution to the dynamic pricing problem is given via Bayesian dynamic programming. These papers focused on investigating if following the optimal policy obtained via dynamic programming was helpful in recovering the unknown demand information. While, such dynamic programming based policies might lead to an optimal pricing scheme, the approaches tend to be intractable in practice.

In the OR/MS community most work on dynamic pricing has focused on the case of known demands, but with other constraints such as inventory constraints. The problem of DP under unknown demand constraints has only gained traction in this community since mid 2000s. ~\cite{aviv2005partially} were one of the first to look at the problem of dynamic pricing with unknown demand function in this community. They assume that the demand function comes from a parametric family of models with unknown parameter values. They then use a Bayesian approach, where a prior distribution is used on the unknown parameters, and the posteriors are obtained via the Bayes theorem. The problem is modeled as an infinite horizon, discounted revenue maximization problem. They then introduce certainty-equivalence heuristic which is an approximate dynamic programming solution to obtain a DP solution.~\cite{araman2009dynamic, farias2010dynamic} used the same setup as in~\cite{aviv2005partially} but introduced different approximate dynamic programming heuristics to provide a solution to the dynamic pricing problem. While Bayesian, approaches such as the ones mentioned above are attractive as they let you model the parameter uncertainties explicitly, they often lead to intractabilities, such as not being able to compute the posterior in closed form. Non-Bayesian approaches offer  computational benefits over Bayesian methods. In the non-Bayesian context~\cite{carvalho2005learning,carvalho2015dynamic} investigate non-myopic policies for dynamic pricing under unknown demand function. They proposed a variant of one-step lookahead policies which instead of maximizing the revenue at the next step, maximizes revenue for the next two steps. They applied this semi-myopic policy to a binomial demand distribution with logit expectation and compared the performance of this policy with other myopic policies. They showed via simulations that one-step lookahead policies could significantly outperform myopic policies. Various other semi-myopic policies have also been investigated in~\cite{carvalho2015dynamic}.

Semi-myopic policies such as one-step lookahead policies in the non-Bayesian context, and approximate Bayesian dynamic programming based solutions provide tractable dynamic pricing policies. However, these approaches lack theoretical guarantees. A lot of recent work in dynamic pricing has been to obtain suboptimal, yet tractable dynamic pricing policies with provable guarantees.
\cite{besbes2009dynamic} take a non-Bayesian approach to dynamic pricing. They use exploration-exploitation style algorithms where in the first phase (exploration), multiple prices are tried out and demand function estimated at these multiple price points, and in the second phase the optimal price is chosen by solving a revenue maximization problem.~\cite{harrison2012bayesian} considered the problem of DP under the assumption that the true model is one of two given parametric models. They then showed that myopic policies that try to maximize the immediate revenue can fail and myopic Bayesian (and also non-Bayesian policies) can lead to convergence to an incorrect model and hence what they term as incomplete learning.~\cite{broder2012dynamic, den2013simultaneously, keskin2014dynamic} look at various parametric models for the demand function and derive exploration-exploitation based policies using maximum likelihood based formulations. Our work is very similar to the above exploration-exploitation style DP approaches that maximize cumulative revenue over a finite horizon.

Finally, we would like to mention the very recent work of~\cite{ferreira2016online} who study the problem of DP of multiple items under finite horizon, with inventory constraints, and  unknown demand function. They provide a Thompson sampling based solution to maximize the revenue for all the items under the given constraints. The proposed algorithms show promising performance on small, synthetic datasets.

Thompson sampling has been used to tackle the problem of exploration-exploitation in other domains such as computational advertising, feed ranking, reinforcement learning. An exhaustive list is out-of-scope of this paper and we refer the interested reader to the excellent tutorial of~\cite{russo2017tutorial}.

\textbf{Notation.}  Vectors and matrices are denoted by bold letters. Given a vector $\vx_t$, $x_{i,t}$ denotes the $i^{\text{th}}$ dimension of $\vx_t$.
\section{Architecture of a dynamic pricing system}
Any commercial dynamic pricing system needs to figure out how to model the demand function and then how to use the modeled demand function to calculate optimal prices to be applied every day. The DP system that we introduce in this paper has different components which all implement statistical learning and convex optimization algorithms to generate prices daily. We shall now look at each of these components in brief.

\textbf{Demand modeling.} A standard approach to demand modeling in pricing problems is to assume that the demand function comes from some parametric family, and then estimate the parameters of the underlying function using statistical techniques. Lot of parametric models have been investigated in pricing and revenue management literature such as linear models, log-linear models, constant elasticity models and logit model~\cite{talluri2006theory}. Parametric modeling has been particularly popular because it allows one to build simple, explainable models and also allows for simple pricing algorithms. However, commonly used models, both parametric and non-parametric, in demand modeling tend to assume that the underlying demand function is stationary. In reality the demand function is not stationary and one needs to model the dynamic nature of the demand function in real world pricing systems. In order to do this we adapt commonly used demand functions to account for possible changes in the underlying demand function. As previously mentioned a commonly used demand function is the constant elasticity model, which models the demand of item $i$ as
\begin{equation}
d_i (p_i) = f_i \left(\frac{p_i}{p_{0,i}}\right)^{\gamstari}
\end{equation}
Here, $d_i(p_i)$ is the demand of item $i$ at price $p_{0,i}$, $f_i$ is the baseline demand at price $p_{0,i}$ for item $i$ and $\gamstari<-1$ is the price elasticity of item $i$. In order to account for possible changes in the underlying demand function we instead model the demand function, at time $t$ for item $i$, using the equation
\begin{equation}
\label{eqn:power}
\dit (p_i) = \doit \left(\frac{p_i}{\poit}\right)^{\gamstari}
\end{equation}
Here $\doit$ is the demand forecast for item $i$ on day $t$ if the price is $\poit$, and $\gamstari$ is the price elasticity of item $i$. Our DP engine has two dedicated components namely demand forecasting component, and elasticity estimation component that estimate $\doit$, $\gamstari$ every day.

\textbf{Demand forecasting component.} The demand forecasting module at time $t$, for item $i$, takes the observed demand for the item $i$ in the past at different price points $((p_{i,1},d_{i,1}),\ldots (p_{i,t-1}, d_{i,t-1}))$ and estimates $f_{i,t}$. The demand forecasting module uses multiple models for forecasting, such as time-series models ARMA, ARIMA, EWMA, AR(k) along with non-time series models such as logistic regression and combines these models to get a new model. The weights using in the combination of the models depends on the past performance of individual models. For the purpose of this paper it is enough to treat this as a black-box and one can assume that the demand forecasts are available for each item on each day.

\textbf{Price elasticity component.} The price elasticity component estimates the elasticity parameter $\gamstari$, for each item $i$. $\gamstari$ captures how the demand of item $i$ changes with its price. If $\pit$ is close to $\poit$ then one can approximate the above exponential form by the following linear equation
\begin{equation}
\label{eqn:linapprox}
\dit(p_i) \approx \doit + \left(p_i - \poit\right)\frac{\doit\gamstari}{\poit}.
\end{equation}
From the above linear approximation, elasticity estimation can be reduced to a linear regression problem which can be solved using either ordinary least squares (OLS) approach or via robust linear squares(RLS). We use RLS for elasticity estimation. Furthermore, instead of looking at all of the past data for an item, one can focus only on the most recent data (say $1$ month or $2$ months) to estimate elasticities. Estimating elasticities is generally much more harder than demand forecast. This is because if an item has only $k$ distinct price points in the last few months then one needs to estimate elasticity for this item using only $k$ data points. It is not uncommon to see $k$ being less than $5$ for a lot of items. Hence, elasticity estimation is plagued by data sparsity issue. In such cases one can mitigate the problem to some extent by using multi-task learning techniques~\cite{caruana1998multitask} to group items with similarly elasticities and estimate the elasticity of these items together.

\textbf{Optimization component.} The last component is an optimization engine that outputs prices of different items. For the sake of convenience, it is natural to put multiple items into a single basket, and impose various constraints that the prices the items in the basket could take. These constraints are typically set by business needs, and include constraints such as minimum and maximum prices the items in the basket can have, bounds on how much the price of an item in the basket could change each day, and some basket-wide level constraints such as minimum margin on the basket. If this item is sold by other e-commerce retailers then it is natural to include the prices of the other e-commerce retailers in the constraint set. These constraints can change from time to time, but at each point in time are known to the optimization engine. For the purpose of this paper it is not necessary for us to focus on the exact geometry of the constraint set, but it suffices to assume that the constraint set is always a feasible, convex set,  and "simple" enough to enable optimization. The revenue of an item $i$ in a basket $\cB$ can be estimated as
\begin{align}
\label{eqn:rev}
\Revit(\pit) &= \pit\times \dit(\pit)\\
&\approx \pit (\doit + \left(\pit - \poit\right)\frac{\doit\gamstari}{\poit})\\
&= \frac{\pit^2\doit\gamstari}{\poit} - \pit\doit\gamstari + \pit\doit
\end{align}
where the first line corresponds to the definition of revenue, and the second line comes from the linear approximation of the demand function as shown in equation~\eqref{eqn:linapprox}. Let, $\p=[p_1,p_2,\ldots, p_{|\cB|}]$ be a vector of prices of items in the basket $\cB$. The optimization engine solves the following optimization problem using estimates for the quantities $\gamstari, \doit$ obtained from the elasticity estimation and demand forecasting module respectively.
\begin{align}
\label{eqn:maxrev}
\pt &= \arg\max_{\p} \sum_{i \in \cB} \frac{\pi^2\doit\gamstari}{\poit} - \pi\doit\gamstari + \pi\doit\\
&\text{subject to: } \p=[p_1,p_2,\ldots, p_{|\cB|}] \in \cCt \nonumber
\end{align}
Since price elasticities are negative, the above optimization problem is maximizing a concave function subject to convex constraints. Hence, it is a convex optimization problem that can be solved using standard techniques~\cite{bertsekas2015convex}. We call the optimization problem in Equation~\eqref{eqn:maxrev} as \maxrev optimization problem as it tries to maximize the revenue of a basket under constraints. When \maxrev is used along with passive elasticity estimation techniques (for example using OLS or RLS as mentioned above) we call the resulting algorithm as \maxrevps. The entire architecture can be summarized in Figure~\eqref{algo:maxrevps}.
\begin{algorithm}[t]
\SetAlgoNoLine
\KwIn{A basket $\cB$, and time period $T$ over which we intend to maximize cumulative revenue}
\For{$t \gets 1$ \textbf{to} $T$} {
\begin{enumerate}
 \item For each item $i \in \cB$ calculate their demand forecasts using the demand forecaster.
  \item For each item $i \in \cB$ calculate their price elasticities $\gamstari$.
  \item Solve the \maxrev optimization problem, shown in Equation~\eqref{eqn:maxrev} to obtain new prices $\pt$.
  \item Apply these prices and observe the revenue obtained $R_t$.
\end{enumerate}
}
\caption{Architecture of the proposed dynamic pricing engine.}
\label{algo:maxrevps}
\end{algorithm}
\section{Towards active algorithms for dynamic pricing}
The \maxrevps algorithm provides a dynamic pricing algorithm that estimates demand forecasts and elasticity passively, and uses these estimates to solve a \maxrev optimization problem. As mentioned in the introduction, dynamic pricing algorithms that use passive learning are incomplete learning algorithms and do not maximize revenue in the long run. From a practical standpoint, estimating demand forecasts and elasticity is not easy. Price elasticity of an item reflects how the demand changes when the price of an item changes. If the price of an item does not change often (a typical case in e-commerce) then it becomes very hard to estimate the price elasticity, which can in turn lead to fewer price changes, and sub-optimal price changes. Moreover, due to the poor estimation of elasticities, there is high uncertainty in elasticity estimates. One approach to factor into account the inaccuracies in elasticities is to model the \maxrev problem as a robust optimization problem~\cite{ben2009robust}. In a robust optimization formulation of \maxrev, one considers the elasticity parameter as an uncertain quantity and tries to maximize revenue w.r.t. the worst possible value that the elasticity can take. For example, the robust \maxrev problem would be
\begin{align*}
\pt &= \arg\max_{\p}\min_{\gamstar \in \cG} \sum_{i \in \cB} \frac{\pi^2\doit\gamstari}{\poit} - \pi\doit\gamstari + \pi\doit\\
&\text{subject to: } \p=[p_1,p_2,\ldots, p_{|\cB|}] \in \cCt \nonumber
\end{align*}
where the set $\cG\subset \bbR^{|\cB|}$ is a set of possible values for $\gamstar$. The robust optimization formulation replaces a point estimate for the vector $\gamstar$ with one of infinite possible values, constrained to be in the set $\cG$. The problem with this approach is that this formulation can be very pessimistic, and  there is no clear way to incorporate feedback to update the set $\cG$ as we gather more data.

An alternative approach to handle uncertainties in the parameters is to consider the \maxrev problem as an optimization problem under uncertainty. This allows us to systematically model the uncertainty in our parameters, and use the feedback from our previous pricing actions to update parameter uncertainties.

\subsection{Decision making under uncertainty}
Decision making under uncertainty is a broad area, which as the name suggest, focuses on how to make decisions to maximize some objective, when there is uncertainty in estimates related to the objective function. This broad class of problems has found various applications such as in search, online advertising, online recommendations, route planning.  For example in search it is not clear what the click-through-rate (CTR) of certain documents are because these documents have never been shown to users or has been shown very few times. Solutions to such problems involves exploration, which means that we make potentially sub-optimal decisions as long as the feedback from these decisions can improve our estimates. While making sub-optimal decisions (exploration) can help gather more information it can also hamper our experience, and therefore one needs to be careful when trying to explore. This problem is known as exploration-exploitation trade-off. Passive learning algorithms such as \maxrevps shown in Figure~\eqref{algo:maxrevps} are focused purely on exploitation and do not explore enough and hence lose out on revenue in the long run. Various algorithms and models have been devised to handle the exploration-exploitation trade-off. The best known model is the multi-armed bandit model (MAB), which can be described as follows: There are $k$ arms, and choosing an arm gives an i.i.d. reward from a fixed unknown probability distribution that depends on the arm. Choosing an arm $i$ gives no information about any other arm $j\neq i$. An agent is tasked with obtaining the maximum possible reward in $T$ rounds, where in each round she chooses one of the $k$ arms and obtains an i.i.d. reward associated with the arm distribution. A simple algorithm is to pull each arm once (or may be a few times each), calculate the average reward of each arm and then for the rest of the time choose the arm which gave the maximum reward.  This simple solution is sub-optimal as it is too greedy and suffers the same problem as the \maxrevps algorithm. Another simple solution is to choose the arm with the best reward each time with probability $1-\epsilon$, but with probability $\epsilon$ choose a random arm. When $\epsilon = 0$, this reduces to a greedy algorithm, when $\epsilon = 1$ this becomes a purely explorative algorithm. However, with the right choice of $\epsilon$, it is possible to do well. Such algorithms are known as $\epsilon$-greedy algorithms. The performance of these algorithms are measured in terms of regret, which tells us how well the algorithm does in comparison to an oracle algorithm that knows the best arm a priori. If the average regret of the algorithm converges to $0$, then the algorithm is said to be consistent. While $\epsilon$-greedy is consistent for the right value of $\epsilon$, better algorithms exist, two of which we cover here in brief. A generalization of the multi-armed bandit problem is the bandit optimization problem where one tries to solve an optimization problem with certain parameters being unknown. For example, in bandit linear optimization problem one tries to solve the optimization problem $\max_{\vx\in \cC} \vx^\top \vtheta_{\star}$, where $\cC$ is a known polytope and $\vtheta_{\star}$ is an unknown vector. The bandit game proceeds in rounds where in each round the learner proposes a $\vx$, and gets to see a noisy evaluation of the objective function $\vx^\top\vtheta_{\star}$. For an extensive survey of various bandit algorithms and bandit models we refer the interested reader to the survey of~\cite{bubeck2012regret}.

The Upper Confidence Bound (UCB) algorithm~\cite{auer2002nonstochastic,auer2002finite} constructs confidence bounds on the reward of each arm, and chooses the arm which has the largest upper confidence bound. The UCB algorithms and its variants are optimal, consistent algorithms. Implementing UCB requires one to accurately estimate confidence bounds, which might not be easy in complex decision making problems.

An alternative to the UCB algorithm is the Thompson Sampling algorithm (TS)~\cite{thompson1933likelihood, gopalan2014thompson, russo2017tutorial}. TS is a Bayesian algorithm, which places a prior distribution on the reward of the arms, and this distribution gets updated, using the Bayes rule, as one gathers feedback from previous actions. The TS algorithm is based on probability matching, i.e. an arm is chosen with probability equal to the probability of the arm being optimal. Like UCB, TS is also a consistent algorithm,  but can be easier to implement than the UCB algorithm, even in complex decision making problems. This is because TS avoids explicitly constructing confidence intervals around the uncertain parameters. These properties of TS make it an attractive choice of an explore-exploit algorithm for real-world applications.
\subsection{Dynamic pricing as a bandit optimization problem and Thompson sampling}
In order to apply TS to our problem of maximizing revenue, we shall reformulate our \maxrev problem as a bandit optimization problem. This is shown in game~\eqref{game:bandit_game}. Our bandit formulation explicitly models the uncertainty in the true elasticity values, but assumes that there is no uncertainty in our demand forecasts. That is, we assume that the demand forecasting module is perfect and spits out accurate forecasts. This assumption is motivated by the fact that demand forecasts are relatively easier to estimate when compared to price elasticities, as they do not suffer from data sparsity problem. For this reason, our bandit formulation of DP focuses on modeling uncertainties in price elasticities.

\begin{game}
\SetAlgoNoLine
\KwIn{A basket, $\cB$, containing $B$ items, and the time period $T$ over which we intend to maximize revenue.}
Oracle chooses a vector $\gamstar\preceq 0$, but does not reveal it to the player\;
\For{$t \gets 1$ \textbf{to} $T$} {
The player plays a price vector $\vp_t$ for the basket $\cB$\;
The player obtains a stochastic reward (revenue) $R_t$, such that $\bbE[R_t]=\Revt(\vp_t;\gamstar, \doot,\ldots \doBt )$ where $\Revt$ is given by Equation~\eqref{eqn:revts} and the demand forecast used in Equation~\eqref{eqn:revts} are assumed to have been calculated without any uncertainty using the demand forecast component of our dynamic pricing system\;
}
\caption{\maxrev problem as a bandit optimization problem}
\label{game:bandit_game}
\end{game}
TS begins by putting in a prior distribution over the unknown parameters. Since, $\gamstar$ is unknown to us, we shall put a prior distribution over $\gamstar$. Since we observe the revenue corresponding to out pricing decisions each day, we put a likelihood model on the observed revenue. Let, $\Rt$ be the observed revenue of a basket~\footnote{We define the revenue of a basket as the sum of revenues of all items.}. Our probabilistic model is given by the following equations
\begin{align}
\Pi_0(\gamstar)&=\cN(\vmu_0,\Sigma_0).\label{eqn:prior}\\
l(\Rt;\Revt,\gamstar)&=\cN(\Rt;\Revt,\sigma^2).\label{eqn:likelihood}\\
\Revt (\vp; \gamstar,\dov_t)& = \sum_{i \in \cB} \frac{\pi^2\doit\gamma_{\star,i}}{\poit} - \pi\doit\gamma_{\star,i} + \pi\doit. \label{eqn:revts}\\
\Pi_{t}(\gamstar)&\propto \Pi_{t-1}(\gamstar)N(\Rt;\Revt,\sigma^2).\label{eqn:update}
\end{align}
Equation~\eqref{eqn:update} provides us a way to obtain the posterior on the parameter $\gamstar$ using Bayes update. In order to make this equation concrete we need to see how $\Revt$ depends on $\gamstar$. Let,
\begin{align}
\vtheta_{ti}&=\frac{\pi^2\doit}{\poit}-\pi\doit.\\
\Rtbar&= \pi\doit.
\end{align}
It is then easy to see that $\Revt = \gamstar^\top\vtheta_t + \Rtbar$.
\begin{align*}
N(\Rt; \Revt,\sigma^2)& = N (\Rt;\gamstar^\top\vtheta_t+\Rtbar,\sigma^2)\\
& \propto \exp\left(-\frac{1}{\sigma^2}(\Rt-\Rtbar - \gamstar^\top\vtheta_t)^2\right)\\
&\propto \exp\left(-(\gamstar-\betat)^\top\mM_t^{-1}(\gamstar-\betat)\right).
\end{align*}
where,
\begin{align*}
\mM_t^{-1}\vbeta_t = \frac{\Rt-\Rtbar}{\sigma^2}\vtheta_t.\\
\mM_t^{-1} = \frac{1}{\sigma^2}\vtheta_t\vtheta_t^\top + \lambda I.
\end{align*}
\begin{align}
\Pi_{t}(\gamstar)&\propto
\cN(\gamstar;\vmu_{t-1},\mSigma_{t-1})\cN(\vgamma;\vbeta_t,\mM_t)\\
&\propto \cN(\gamstar;\vmu_{t},\mSigma_{t}),
\end{align}
where, $\vmu_{t},\mSigma_{t}$ are given by the following expressions
\begin{align}
\label{eqn:update_mu}
\vmu_{t}&=(\mSigma_{t-1}^{-1}+\mM_t^{-1})^{-1}\left(\mSigma_{t-1}^{-1}\vmu_{t-1}+\frac{R_t-\bar{R}_t}{\sigma^2}\vtheta_t\right).\\
\mSigma_{t}&=(\mSigma_{t-1}^{-1}+\mM_t^{-1})^{-1}\label{eqn:update_sigma}.
\end{align}
The expression for $\mM_t$ is calculated as follows
\begin{align}
\mM_t^{-1}&=\frac{\vtheta_t\vtheta_t^\top}{\sigma^2}+\lambda I.\\
\mM_t^{-1}\beta_t&=\frac{(R_t-\bar{R}_t)\theta_t}{\sigma^2}.
\end{align}
$\lambda>0$ is a small positive constant that has been added to the matrix $\mM_t$ to make it invertible. Equation~\eqref{eqn:prior} puts a Gaussian prior on the elasticity vector $\gamstar$
and equation~\eqref{eqn:likelihood} puts a Gaussian likelihood on the observed revenue. One might, argue that since elasticity parameter is negative, and revenue is always positive a Gaussian prior and Gaussian likelihood are incorrect models. While this criticism is justified, we argue that using this combination of prior and likelihood leads to very simple posterior updates. Moreover, in our TS implementation, we perform rejection sampling on the elasticity parameter. We reject elasticity samples that are positive, which alleviates the problem of the prior having non-zero mass on elasticity values that are positive. One might put more appropriate distributions on both elasticity and revenue which reflect the fact that these parameters take only a restricted set of values (e.g. a log-normal distribution might be more appropriate). However, with such distributions it is generally hard to obtain closed form updates and one has to resort to approximate sampling techniques~\cite{kawale2015efficient,gopalan2014thompson, osband2015bootstrapped, blei2017variational}. Use of such techniques for scalable, active dynamic pricing will be investigated in future work. The pseudo-code for \maxrevts is shown in algorithm~\eqref{algo:maxrevts}.
\begin{algorithm}
\SetAlgoNoLine
\KwIn{A basket, $\cB$, containing $B$ items, and a time period $T$ over which revenue needs to be maximized.}
Oracle chooses a vector $\gamstar\preceq 0$, but does not reveal it to the player\;
Initialize the prior $\Pi_0(\gamstar)$\;
\For{$t \gets 1$ \textbf{to} $T$} {
Keep sampling from $\vgamma_t\sim \Pi_{t-1}$ until all the components of $\vgamma_t$ are negative\;
Use the demand forecaster to obtain demand forecasts $\doit$ for all $i \in \cB$\;
Solve the \maxrev optimization problem shown in Equation~\eqref{eqn:maxrev}, with $\doit,\poit$, and $\vgamma_t$ as an estimate for $\gamstar$, to get price vector $\vp_t$\;
Apply the prices $\vp_t$ to obtain reward $\Rt$ \;
Perform the updates in \eqref{eqn:update_mu} - \eqref{eqn:update_sigma} to get updated distribution $\Pi_t(\gamstar)$.
}
\caption{\sc \maxrevts - Active learning based dynamic pricing algorithm}
\label{algo:maxrevts}
\end{algorithm}
\subsubsection{\textbf{Choice of prior and scalability of updates}}
The update equations used by \maxrevts algorithm needs a prior distribution, and an estimate for the noise variance $\sigma^2$. It is common to have prior estimates for elasticities using historical data. These prior estimates can be used as the mean $\vmu_0$ of the prior distribution. We set $\Sigma_0 = cI$, for some appropriate constant $c$. $c$ should be large enough to perform exploration, but not too large enough to wash out values in $\vmu_0$. An estimate of $\sigma$ can be obtained using historical data to calculate the sample standard deviation of the observed revenue of the basket. Finally, by keeping only a diagonal approximation of the covariance matrix and using Sherman-Morrison-Woodbury identity we can get away with $O(|\cB|)$ runtime and storage complexity.
\section{Experiments}
We shall now demonstrate the performance of \maxrevts and \maxrevps on both synthetic and real-world datasets.
\subsection{Results on a synthetic dataset}
\label{sec:synth_expts}
\begin{figure}
\includegraphics[scale = 0.3]{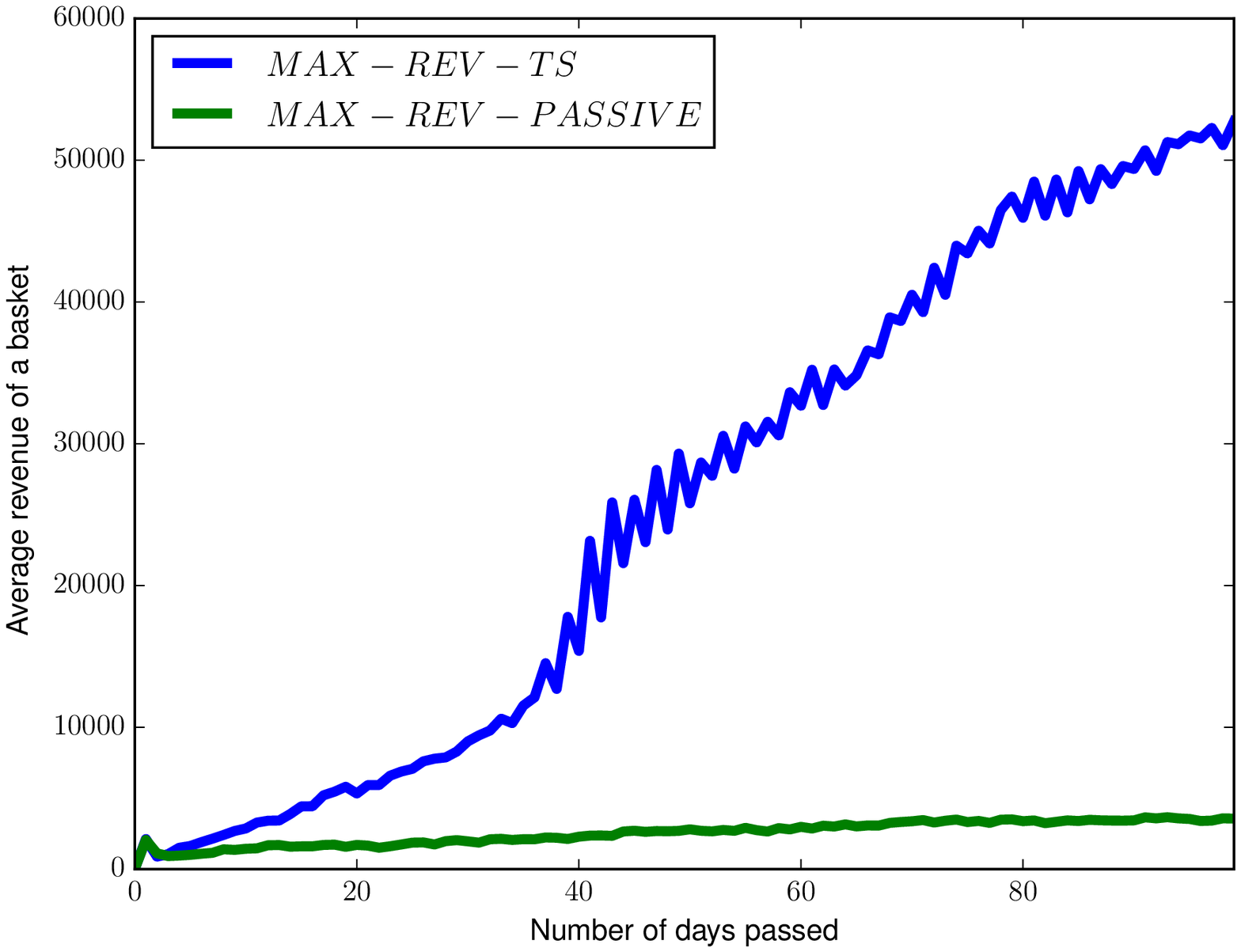}
\caption{\label{fig:synthexpts} Plot of the average revenue of a basket (averaged over 10 trials) w.r.t. time for both $\textsc\maxrevts$ and $\textsc\maxrevps$ algorithms. $\textsc{\maxrevts}$ improves its estimates and  better optimizes for the revenue of the basket when compared to $\textsc{\maxrevps}$}
\end{figure}

We generated synthetic datasets for testing the performance of \maxrevts and \maxrevps. In order to do that, we create a basket of $100$ items, each with elasticity in the interval $[-3,-1]$. This interval was chosen because in our real-world experiments we noticed that most items have elasticities in this range. At time $t$ for each item, we generate a demand forecast for the item $\doit$ using an auto regressive process on the demands previously observed, and use the exponential formula
$\dit(\pit) = \max(\doit \left(\frac{\pit}{\poit}\right)^{\vgamma_{\star,i}}+\epsilon_t,0)$ to generate the demand at the price $\pit$. All prices are constrained to belong to the set $\cCt = [10,20]$. This is repeated for $T = 100$ rounds for both \maxrevts and \maxrevps. Notice that since \maxrevts and \maxrevps have different behaviours, the prices that they generate could be different and hence the datasets that they generate can be very different. Nevertheless, both the algorithms generate datasets using the same formula for the demand forecast, demand and elasticities, and have the same values for $\doio, p_{i,0}$. The data generating mechanism used by \maxrevts and \maxrevps is shown in Algorithm~\eqref{algo:syndata}
\begin{algorithm}
\SetAlgoNoLine
\KwIn{A basket, $\cB$, containing $B = 50$ items, a small constant $c_0>0$, and an algorithm $\cA$ (such as \maxrevts or \maxrevps)}
Choose a vector $\gamstar\in [-3,-1]^{100}$, but do not reveal it to the algorithm.\;
Set $\beta = 0.5$, $p_{i,0} = 12$ for all $i \in \cB$, $\doio \in [0.5,5]$ for each item.\;
\For{$t \gets 1$ \textbf{to} $T$} {
Calculate the demand forecast for each item $i$ using the formula $\doit = c_0 + \sum_{\tau = t-1}^0\beta^{t-\tau}d_{i,\tau} + \epsilon_t$\;
Use $\doit, \vgamma_{\star,i}$, and the price $\pit$ generated by algorithm $\cA$, and the exponential formula $\dit(\pit) = \max\left(\doit \left(\frac{\pit}{\poit}\right)^{\vgamma_{\star,i}} + \epsilon_t, 0\right)$ to generate the demand $d_i(\pit)$ of the item $i$\;
Register $(\doit,\poit,\dit)$ as a data point generated by the algorithm $\cA$.
}
\caption{Data generating mechanism for synthetic experiments}
\label{algo:syndata}
\end{algorithm}
We used the above data generating mechanism to generate synthetic datasets for both \maxrevts and \maxrevps. The random vectors $\gamstar$ and $\doio$ are generated once and fixed for both the algorithms. This way we can guarantee that the starting point for both the algorithms is the same. The random noise added to demand forecast estimation and the demand are sampled independently from $N(0,1)$ distribution. We are interested in calculating the total revenue  of all items in the basket on each day. We ran $10$ independent trials for each algorithm and report the average of the total revenue for both the algorithms in Figure~\eqref{fig:synthexpts}. As one can see from this figure, $\sc{\maxrevts}$ continues to learn and obtains increasing revenue as time progresses. In contrast, $\sc{\maxrevps}$ learns very slowly and the average revenue stays almost the same with time.
\subsection{Results on a real-world dataset}
We now show our results on a real world e-commerce dataset. While applying the \maxrevts algorithm in the case of synthetic experiments as shown in Section~\eqref{sec:synth_expts} is fairly straightforward, in a real world experiment with real e-commerce transactions there are many more complications that we need to handle carefully. We would like to discuss a few such issues that we encountered during our implementation.
\begin{enumerate}
\item In our discourse we considered the basket to be static. In reality basket changes from time to time. This is because items get added or removed from basket by merchants as per business needs.
\item On certain items, due to business constraints, the prices get fixed by business. For example, consumables such as diapers, cereal are high visibility items and their prices are almost always fixed. Items on which a merchant runs promotions also have their prices fixed. Items can go out-of-stock, unpublished from the site, or may be retired. We mark all such items and do not apply TS on those items.
\item Both \maxrevts and \maxrevps rely on the constant elasticity model to estimate the demand function. This model needs $\doit$ at each point of time. The demand forecaster estimates this quantity. For items that have not had sales in the recent past, the demand forecaster can predict $\doit = 0$. If $\doit = 0 $, then our demand function becomes a $0$ function, i.e. $\dit(p) = 0$, which makes the revenue function a constant and "shorts" our \maxrev optimization problem and we lose information when $\doit=0$. In general, pricing is much more harder when $\doit \approx 0$. For this reason, we do not apply TS on items with a very low demand forecast.
\item Feedback in a real-world e-commerce system is delayed. This is because when we apply new prices on items, one needs to wait for at least one day to see how the market responded to new prices, and what the revenue was at the new prices.
\end{enumerate}
\subsection{Experimental setup}
\label{sec:expt_setup}
\begin{figure}
\includegraphics[scale = 0.3]{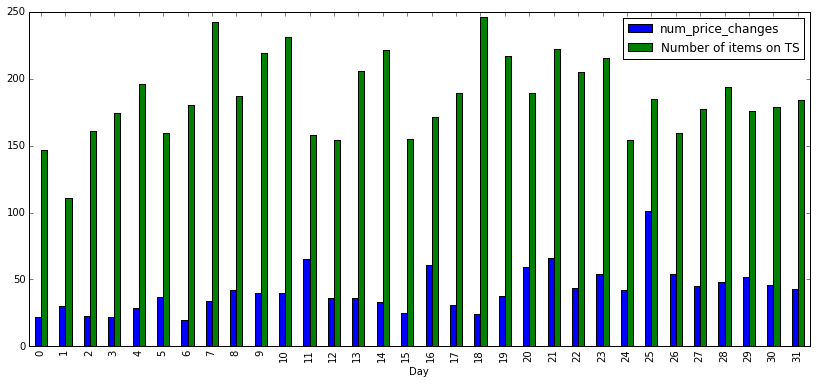}
\caption{\label{fig:price_changes_ts} The total number of items in baskets $\Bo, \Bt$ which are on TS each day and the number of items whose price changed compared to the previous day.}
\end{figure}

We experiment on two separate, unrelated baskets. The first basket $\Bo$ has roughly 19000 items and the second basket, $\Bt$ has roughly 5000 items. Each of these baskets have well defined parameters such as margin goals that need to be achieved on the basket, upper and lower bounds on the price of the items. Each basket $\cB$ can be partitioned into two disjoint sets, namely $\Bf, \Bm$. The part $\Bf$ consists of items whose prices are already pre-fixed by various business constraints and hence they are not part of the $\maxrev$ optimization problem. $\Bm$ is the subset of the basket on which we run the \maxrev solver that generates a new set of prices. This subset consists of items that are published in-stock, with no active promotions. The set of items in $\Bm$ can be further broken down into two disjoint part, namely $\Bp,\Bts$. $\Bp\subset \Bm$ is the set of items which are not eligible for TS, because of low demand forecast. We use a demand forecast threshold of $2$. For items in $\Bp$ we use passive elasticity estimation methods such as OLS/RLS to calculate their elasticities. We apply the \maxrevts algorithm on the set $\Bts$. A caveat of our approach is that the \maxrev optimization problem solved by \maxrevts includes not just items in the partition $\Bts$ but also items in the partition
$\Bp$. Hence, our implementation is only an approximate $\maxrevts$ algorithm. One may avoid such a partition based approach and instead apply TS on all items in $\Bm$. However, in our experience such implementations have very poor performance as they include items with low demand forecasts, for which our constant elasticity demand models are inaccurate.  We ran $\maxrevts$ for about five weeks. On every third day, TS updates were performed to get an updated posterior distribution over their elasticities.  The number of items that are on TS, on a particular day, in baskets $\Bo, \Bt$ combined is shown by the green bar in Figure~\eqref{fig:price_changes_ts}.
By design, TS encourages exploration of elasticity values for different items. This exploration in elasticity space also forces exploration of the price-demand curve of an item. Hence, we expect a good number of price changes among items that are put on TS.  The blue bar in Figure~\eqref{fig:price_changes_ts} tells us among the items that were put on TS, for how many items did we observe a price change, compared to their prices on the previous day. As we can see from this figure, that on roughly $20\% - 30\%$ of the items there were price changes.

\subsection{Revenue contributions before the start of TS and after TS was applied.}
\begin{figure}
\includegraphics[scale = 0.3]{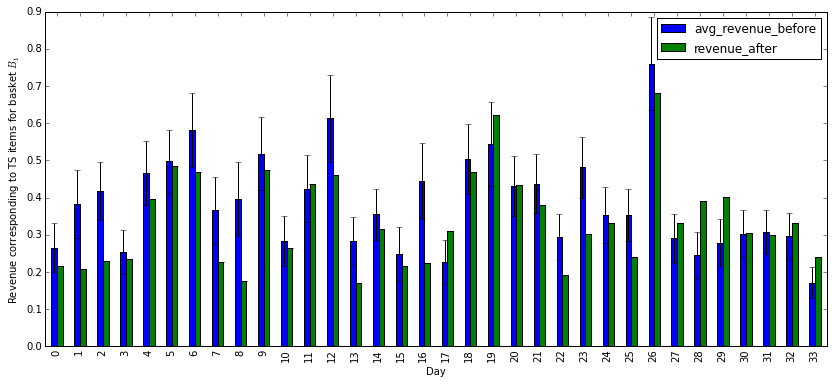}
\includegraphics[scale = 0.3]{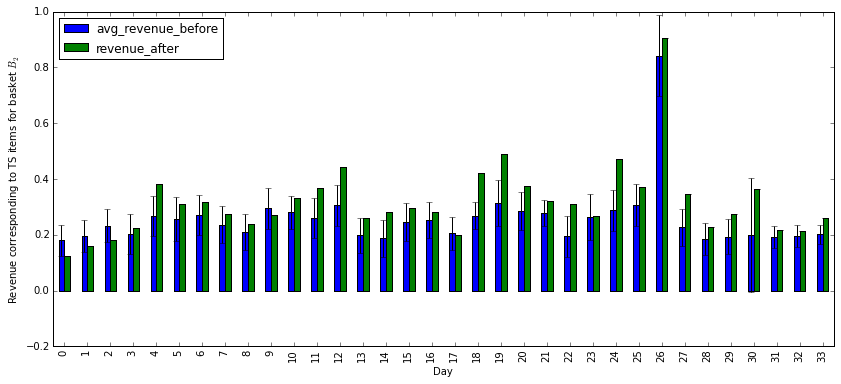}
\caption{\label{fig:before_after_ts} The green bar corresponds to the revenue contributed by items that were eligible for TS on that day. The blue bar corresponds to the revenue of these items, averaged over a period of $30$ days before the start of the TS period. We also have shown the standard deviation of the revenue before the start of TS period. All quantities in this plot have been normalized.}
\end{figure}

In Figure~\eqref{fig:before_after_ts} we look at the revenue obtained by items that were put on TS each day. For items put on TS, each day, we also calculate what was the average revenue (and the standard deviation) corresponding to these items before the start of TS experiment. This average was calculated over a span of $30$ days before the start of TS period. Since, \maxrevps was applied before the start of TS period, this results provide a comparison of \maxrevts with \maxrevps. On basket $\Bo$, from a revenue point of view initially TS does not seem to help as it gets smaller revenue as compared to not applying TS. This is understandable, since initially the variance in our estimates of elasticity is large. However, about half-way through the experimental period the revenue contribution due to items on TS improves and is competent with the average revenue before the start of TS period.
On basket $\Bt$, the results suggest that TS consistently outperforms the revenue obtained before the start of TS period.
\subsection{Statistical significance tests.} Figure~\eqref{fig:before_after_ts} compares the revenue of items that were put on TS each day, with the average revenue of these same set of items before the start of TS. We now present quantitative results that looks at all items that were on TS at least some number (say $k$) of times. For these items we calculate $\delta_i$, which is the difference between the average revenue of item $i$ on days when it was on TS and the average revenue for a period of $30$ days before the start of TS. For each value of $k$, let $S_k$ be the number of items that were on TS at least $k$ times during the TS period.

For each $k$ and for each basket we perform the following hypothesis test. $H_0: \bbE[\delta_i] = 0$, $H_1: \bbE[\delta_i] \neq 0$. We use Wald's test~\cite{wasserman2013all} to perform this hypothesis test and report significance at level $\alpha = 0.05$.

Table~\eqref{tab:hyp_test_B1} suggests that for basket $\Bo$, barring $k = 30$, all the p-values of the Wald test are less than $\alpha = 0.05$. This means that except for $k = 30$ the null hypothesis can be rejected. Moreover the last column of the table shows that $\bbE[\delta_i]<0$, which means that there is a statistically significant degradation in the revenue per item, for basket $\Bo$, when the revenue is calculated only over those days where TS was applied. In contrast to this negative result, Table~\eqref{tab:hyp_test_B2} suggests that barring the case of $k=1$, there is a statistically significant increase in revenue per item, for items in basket $\Bt$, due to the application of TS.
\begin{table}
\begin{tabular}{|l|l|l|l|}
\hline
k&$S_k$& p-value & mean of $\delta_i$\\\hline
5&82&0.039&-14.19\\\hline
10&55&0.016&-13.35\\\hline
15&26&0.061&-17.67\\\hline
20&20&0.027&-25.04\\\hline
25&15&0.023&-33.08\\\hline
30&8&0.29&-21.58\\\hline
\end{tabular}
\caption{\label{tab:hyp_test_B1}Wald's test results for basket $\Bo$ comparing the difference in revenues due to TS. Let $\delta_i$ be the difference between the average revenue of item $i$ on the days this item was put on TS, and the average revenue of this item over a period of $30$ days before the start of TS. $k$ in the table filters out items which have been on TS for less than $k$ days. The last column is the sample average of $\delta_i$.}
\vspace{-10pt}
\end{table}
\begin{table}
\begin{tabular}{|l|l|l|l|}
\hline
k&$S_k$& p-value & mean of $\delta_i$\\\hline
5&371&$2.37x10^{-6}$&1.92\\\hline
10&197&$1.89x10^{-7}$&2.08\\\hline
15&128&$2.13x10^{-6}$&2.68\\\hline
20&91&0.000628&1.966\\\hline
25&67&0.00109&2.21\\\hline
30&44&0.0019&3.062\\\hline
\end{tabular}
\caption{\label{tab:hyp_test_B2} Wald's test results for basket $\Bt$ with the same setup as in Table~\eqref{tab:hyp_test_B1}.}
\vspace{-10pt}
\end{table}
In the above hypothesis testing experiments we reported results of the revenue difference per item when calculated only over the days when the items were eligible for TS. We now calculate a different metric which is similar in spirit to the one calculated in tables~\eqref{tab:hyp_test_B1},~\eqref{tab:hyp_test_B2}, where instead of calculating the revenue of an item averaged only over days when the item was on TS, we now average the revenue of the item during the entire TS period. For each item $i$ we calculate $\delta_i$ which is the difference in the average revenue of item $i$ after the start of TS, and the average revenue of the item in the $30$ day window before the start of TS. The null hypothesis for our test is that $H_0: \bbE \delta_i =0$  and the alternate hypothesis is $H_1: \bbE \delta_i \neq 0$. We use Wald's test as our testing procedure and in tables~\eqref{tab:hyp_test_B1_all}, ~\eqref{tab:hyp_test_B2_all} report the p-value of the test and also the sample average of $\delta_i$ calculated over all qualifying items. As one can see from table~\eqref{tab:hyp_test_B1_all}, for basket $\Bo$ there is a degradation in the revenue per item, though not statistically significant. In contrast, for basket $\Bt$, as seen from Table~\eqref{tab:hyp_test_B2_all}, there is a statistically significant increase in the average revenue per item after the application of TS.
\begin{table}
\begin{tabular}{|l|l|l|l|}
\hline
k&$S_k$& p-value & mean of $\delta_i$\\\hline
5&82&0.129&-9.65\\\hline
10&55&0.328&-4.28\\\hline
15&26&0.424&-5.45\\\hline
20&20&0.25&-10.19\\\hline
25&15&0.19&-14.27\\\hline
30&8&0.52&-12.76\\\hline
\end{tabular}
\caption{\label{tab:hyp_test_B1_all} Wald's test results for basket $\Bo$ comparing the difference in revenues due to TS. Let $\delta_i$ be the difference between the average revenue of item $i$ during the TS period (irrespective of whether $i$ was on TS or not on a certain day), and the average revenue of this item over a period of $30$ days before the start of TS. $k$ in the table filters out items which have been on TS for less than $k$ days. The last column is the sample average of $\delta_i$.}
\vspace{-10pt}
\end{table}
\begin{table}
\begin{tabular}{|l|l|l|l|}
\hline
k&$S_k$& p-value & mean of $\delta_i$\\\hline
5&371&$2.28x10^{-8}$& 1.582\\\hline
10&197&$3.13x10^{-7}$&1.649\\\hline
15&128&$9.64x10^{-6}$&2.05\\\hline
20&91&0.00099&1.64\\\hline
25&67&0.00084&1.99\\\hline
30&44&0.0013&2.74\\\hline
\end{tabular}
\caption{\label{tab:hyp_test_B2_all}Wald's test results for basket $\Bt$ with the same setup as in Table~\eqref{tab:hyp_test_B1_all}.}
\vspace{-20pt}
\end{table}
\begin{table}
\begin{tabular}{|l|l|l|l|l|l|}
\hline
Basket&$|\cB|$&$|\Bf|$& $|\Bp|$ & $|\Bts|$ & $\frac{|\Bts|}{|\Bm|}$ \\\hline
$\Bo$&19K&13K&6K&25& 0.004\\\hline
$\Bt$&5K&3.3K&1.5K&150& 0.1\\\hline
\end{tabular}
\caption{\label{tab:sizes} Approximate size of the problem for different baskets. $|\cB|$ is the size of the basket, $|\Bf|$ is the number of items whose prices are pre-fixed by business constraints, $|\Bp|$ is the number of items on which the \maxrev optimization problem is applied but whose elasticities are generated using passive algorithms, and $|\Bts|$ is the number of items on which the \maxrev optimization problem is applied but whose elasticities are generated using TS.}
\vspace{-15pt}
\end{table}

\subsection{Discussion of our experimental results}
An interesting observation from our experimental results is that while on basket $\Bo$ there is a decrease in revenue per item that is put on TS, this result is either not statistically significant at $\alpha = 0.05$ or its p-value is large. In contrast, for basket $\Bt$, the increase in revenue per-item is large and statistically significant, and the p-values are small. One may ask, why is there a difference in performance of the \maxrevts algorithm on these baskets? We present two reasons for this. As mentioned in subsection~\eqref{sec:expt_setup} our implementation applies TS only to a subset of the basket, $\Bts$, and the \maxrevts algorithm solves the \maxrev problem over a larger subset $\Bm$. Hence, our implementation of the \maxrevts algorithm is only an approximation implementation. If $|\Bts| << |\Bm|$, then our implementation can be very crude and the TS updates will be very noisy. Hence, as the ratio $\frac{|\Bts|}{|\Bm|}$ decreases the updates of TS will become less accurate. This ratio can be seen as the singal-to-noise ratio of the problem. Secondly, even if TS updates were accurate enough, the \maxrev solver, being optimized over $\Bm$, might trade-off the revenue of $\Bp$ against $\Bts$, if it lead to a larger revenue for $\Bm$ under the provided constraints. Hence, it could be the case that even though we observe a degradation in revenue over the subset $\Bts$, the overall revenue of items $\Bm$ might not suffer. Table~\eqref{tab:sizes} shows the various sizes for baskets $\Bo, \Bt$. The ratio $\frac{|\Bts|}{|\Bm|}$ is $0.004$ for basket $\Bo$, whereas it is $0.1$ for basket $\Bt$, which explains the poor performance of TS on basket $\Bo$.

\section{Conclusions and future work}
In this paper we designed and implemented dynamic pricing algorithms that maximize a reward function under uncertainty. To the best of our knowledge this is the first real world study of active dynamic pricing algorithms in an e-commerce setting. We demonstrated the performance of our algorithms comparing against passive dynamic pricing algorithms that are most commonly used in practice. We see a statistically significant improvement in the per-item revenue when the number of items on which we apply TS is a large fraction of the basket of items. This ratio can be thought of as the SNR of the problem. On a basket, where TS led to a degradation of per-item revenue, we posit that this could be because the SNR was very low. Our future work includes investigations of various other reward functions such as volumes of sales, and experimental study on more baskets where the SNR ratio is large enough to enable reliable measurement of the performance of active dynamic pricing algorithms.

\bibliographystyle{icml2016}
\bibliography{refs}


\begin{thebibliography}{31}


\ifx \showCODEN    \undefined \def \showCODEN     #1{\unskip}     \fi
\ifx \showDOI      \undefined \def \showDOI       #1{#1}\fi
\ifx \showISBNx    \undefined \def \showISBNx     #1{\unskip}     \fi
\ifx \showISBNxiii \undefined \def \showISBNxiii  #1{\unskip}     \fi
\ifx \showISSN     \undefined \def \showISSN      #1{\unskip}     \fi
\ifx \showLCCN     \undefined \def \showLCCN      #1{\unskip}     \fi
\ifx \shownote     \undefined \def \shownote      #1{#1}          \fi
\ifx \showarticletitle \undefined \def \showarticletitle #1{#1}   \fi
\ifx \showURL      \undefined \def \showURL       {\relax}        \fi
\providecommand\bibfield[2]{#2}
\providecommand\bibinfo[2]{#2}
\providecommand\natexlab[1]{#1}
\providecommand\showeprint[2][]{arXiv:#2}

\bibitem[\protect\citeauthoryear{Agarwal, Bird, Cozowicz, Hoang, Langford, Lee,
  Li, Melamed, Oshri, Ribas, et~al\mbox{.}}{Agarwal et~al\mbox{.}}{2016}]%
        {agarwal2016multiworld}
\bibfield{author}{\bibinfo{person}{Alekh Agarwal}, \bibinfo{person}{Sarah
  Bird}, \bibinfo{person}{Markus Cozowicz}, \bibinfo{person}{Luong Hoang},
  \bibinfo{person}{John Langford}, \bibinfo{person}{Stephen Lee},
  \bibinfo{person}{Jiaji Li}, \bibinfo{person}{Dan Melamed},
  \bibinfo{person}{Gal Oshri}, \bibinfo{person}{Oswaldo Ribas},
  {et~al\mbox{.}}} \bibinfo{year}{2016}\natexlab{}.
\newblock \showarticletitle{A multiworld testing decision service}.
\newblock \bibinfo{journal}{{\em arXiv preprint arXiv:1606.03966\/}}
  (\bibinfo{year}{2016}).
\newblock


\bibitem[\protect\citeauthoryear{Aghion, Bolton, Harris, and Jullien}{Aghion
  et~al\mbox{.}}{1991}]%
        {aghion1991optimal}
\bibfield{author}{\bibinfo{person}{Philippe Aghion}, \bibinfo{person}{Patrick
  Bolton}, \bibinfo{person}{Christopher Harris}, {and} \bibinfo{person}{Bruno
  Jullien}.} \bibinfo{year}{1991}\natexlab{}.
\newblock \showarticletitle{Optimal learning by experimentation}.
\newblock \bibinfo{journal}{{\em The review of economic studies\/}}
  \bibinfo{volume}{58}, \bibinfo{number}{4} (\bibinfo{year}{1991}),
  \bibinfo{pages}{621--654}.
\newblock


\bibitem[\protect\citeauthoryear{Araman and Caldentey}{Araman and
  Caldentey}{2009}]%
        {araman2009dynamic}
\bibfield{author}{\bibinfo{person}{Victor~F Araman} {and}
  \bibinfo{person}{Ren{\'e} Caldentey}.} \bibinfo{year}{2009}\natexlab{}.
\newblock \showarticletitle{Dynamic pricing for nonperishable products with
  demand learning}.
\newblock \bibinfo{journal}{{\em Operations Research\/}} \bibinfo{volume}{57},
  \bibinfo{number}{5} (\bibinfo{year}{2009}), \bibinfo{pages}{1169--1188}.
\newblock


\bibitem[\protect\citeauthoryear{Auer, Cesa-Bianchi, and Fischer}{Auer
  et~al\mbox{.}}{2002a}]%
        {auer2002finite}
\bibfield{author}{\bibinfo{person}{Peter Auer}, \bibinfo{person}{Nicolo
  Cesa-Bianchi}, {and} \bibinfo{person}{Paul Fischer}.}
  \bibinfo{year}{2002}\natexlab{a}.
\newblock \showarticletitle{Finite-time analysis of the multiarmed bandit
  problem}.
\newblock \bibinfo{journal}{{\em Machine learning\/}} \bibinfo{volume}{47},
  \bibinfo{number}{2-3} (\bibinfo{year}{2002}), \bibinfo{pages}{235--256}.
\newblock


\bibitem[\protect\citeauthoryear{Auer, Cesa-Bianchi, Freund, and Schapire}{Auer
  et~al\mbox{.}}{2002b}]%
        {auer2002nonstochastic}
\bibfield{author}{\bibinfo{person}{Peter Auer}, \bibinfo{person}{Nicolo
  Cesa-Bianchi}, \bibinfo{person}{Yoav Freund}, {and} \bibinfo{person}{Robert~E
  Schapire}.} \bibinfo{year}{2002}\natexlab{b}.
\newblock \showarticletitle{The nonstochastic multiarmed bandit problem}.
\newblock \bibinfo{journal}{{\em SIAM journal on computing\/}}
  \bibinfo{volume}{32}, \bibinfo{number}{1} (\bibinfo{year}{2002}),
  \bibinfo{pages}{48--77}.
\newblock


\bibitem[\protect\citeauthoryear{Aviv and Pazgal}{Aviv and Pazgal}{2005}]%
        {aviv2005partially}
\bibfield{author}{\bibinfo{person}{Yossi Aviv} {and} \bibinfo{person}{Amit
  Pazgal}.} \bibinfo{year}{2005}\natexlab{}.
\newblock \showarticletitle{A partially observed Markov decision process for
  dynamic pricing}.
\newblock \bibinfo{journal}{{\em Management Science\/}} \bibinfo{volume}{51},
  \bibinfo{number}{9} (\bibinfo{year}{2005}), \bibinfo{pages}{1400--1416}.
\newblock


\bibitem[\protect\citeauthoryear{Ben-Tal, El~Ghaoui, and Nemirovski}{Ben-Tal
  et~al\mbox{.}}{2009}]%
        {ben2009robust}
\bibfield{author}{\bibinfo{person}{Aharon Ben-Tal}, \bibinfo{person}{Laurent
  El~Ghaoui}, {and} \bibinfo{person}{Arkadi Nemirovski}.}
  \bibinfo{year}{2009}\natexlab{}.
\newblock \bibinfo{booktitle}{{\em Robust optimization}}.
\newblock \bibinfo{publisher}{Princeton University Press}.
\newblock


\bibitem[\protect\citeauthoryear{Bertsekas and Scientific}{Bertsekas and
  Scientific}{2015}]%
        {bertsekas2015convex}
\bibfield{author}{\bibinfo{person}{Dimitri~P Bertsekas} {and}
  \bibinfo{person}{Athena Scientific}.} \bibinfo{year}{2015}\natexlab{}.
\newblock \bibinfo{booktitle}{{\em Convex optimization algorithms}}.
\newblock \bibinfo{publisher}{Athena Scientific Belmont}.
\newblock


\bibitem[\protect\citeauthoryear{Besbes and Zeevi}{Besbes and Zeevi}{2009}]%
        {besbes2009dynamic}
\bibfield{author}{\bibinfo{person}{Omar Besbes} {and} \bibinfo{person}{Assaf
  Zeevi}.} \bibinfo{year}{2009}\natexlab{}.
\newblock \showarticletitle{Dynamic pricing without knowing the demand
  function: Risk bounds and near-optimal algorithms}.
\newblock \bibinfo{journal}{{\em Operations Research\/}} \bibinfo{volume}{57},
  \bibinfo{number}{6} (\bibinfo{year}{2009}), \bibinfo{pages}{1407--1420}.
\newblock


\bibitem[\protect\citeauthoryear{Blei, Kucukelbir, and McAuliffe}{Blei
  et~al\mbox{.}}{2017}]%
        {blei2017variational}
\bibfield{author}{\bibinfo{person}{David~M Blei}, \bibinfo{person}{Alp
  Kucukelbir}, {and} \bibinfo{person}{Jon~D McAuliffe}.}
  \bibinfo{year}{2017}\natexlab{}.
\newblock \showarticletitle{Variational inference: A review for statisticians}.
\newblock \bibinfo{journal}{{\it J. Amer. Statist. Assoc.}}
  \bibinfo{number}{just-accepted} (\bibinfo{year}{2017}).
\newblock


\bibitem[\protect\citeauthoryear{Broder and Rusmevichientong}{Broder and
  Rusmevichientong}{2012}]%
        {broder2012dynamic}
\bibfield{author}{\bibinfo{person}{Josef Broder} {and} \bibinfo{person}{Paat
  Rusmevichientong}.} \bibinfo{year}{2012}\natexlab{}.
\newblock \showarticletitle{Dynamic pricing under a general parametric choice
  model}.
\newblock \bibinfo{journal}{{\em Operations Research\/}} \bibinfo{volume}{60},
  \bibinfo{number}{4} (\bibinfo{year}{2012}), \bibinfo{pages}{965--980}.
\newblock


\bibitem[\protect\citeauthoryear{Bubeck, Cesa-Bianchi, et~al\mbox{.}}{Bubeck
  et~al\mbox{.}}{2012}]%
        {bubeck2012regret}
\bibfield{author}{\bibinfo{person}{S{\'e}bastien Bubeck},
  \bibinfo{person}{Nicolo Cesa-Bianchi}, {et~al\mbox{.}}}
  \bibinfo{year}{2012}\natexlab{}.
\newblock \showarticletitle{Regret analysis of stochastic and nonstochastic
  multi-armed bandit problems}.
\newblock \bibinfo{journal}{{\em Foundations and Trends{\textregistered} in
  Machine Learning\/}} \bibinfo{volume}{5}, \bibinfo{number}{1}
  (\bibinfo{year}{2012}), \bibinfo{pages}{1--122}.
\newblock


\bibitem[\protect\citeauthoryear{Caruana}{Caruana}{1998}]%
        {caruana1998multitask}
\bibfield{author}{\bibinfo{person}{Rich Caruana}.}
  \bibinfo{year}{1998}\natexlab{}.
\newblock \showarticletitle{Multitask learning}.
\newblock In \bibinfo{booktitle}{{\em Learning to learn}}.
  \bibinfo{publisher}{Springer}, \bibinfo{pages}{95--133}.
\newblock


\bibitem[\protect\citeauthoryear{Carvalho and Puterman}{Carvalho and
  Puterman}{2005a}]%
        {carvalho2015dynamic}
\bibfield{author}{\bibinfo{person}{Alexandre~X Carvalho} {and}
  \bibinfo{person}{Martin~L Puterman}.} \bibinfo{year}{2005}\natexlab{a}.
\newblock \showarticletitle{Dynamic optimization and learning: How should a
  manager set prices when the demand function is unknown?}
\newblock  (\bibinfo{year}{2005}).
\newblock


\bibitem[\protect\citeauthoryear{Carvalho and Puterman}{Carvalho and
  Puterman}{2005b}]%
        {carvalho2005learning}
\bibfield{author}{\bibinfo{person}{Alexandre~X Carvalho} {and}
  \bibinfo{person}{Martin~L Puterman}.} \bibinfo{year}{2005}\natexlab{b}.
\newblock \showarticletitle{Learning and pricing in an internet environment
  with binomial demands}.
\newblock \bibinfo{journal}{{\em Journal of Revenue and Pricing Management\/}}
  \bibinfo{volume}{3}, \bibinfo{number}{4} (\bibinfo{year}{2005}),
  \bibinfo{pages}{320--336}.
\newblock


\bibitem[\protect\citeauthoryear{den Boer}{den Boer}{2015}]%
        {den2015dynamic}
\bibfield{author}{\bibinfo{person}{Arnoud~V den Boer}.}
  \bibinfo{year}{2015}\natexlab{}.
\newblock \showarticletitle{Dynamic pricing and learning: historical origins,
  current research, and new directions}.
\newblock \bibinfo{journal}{{\em Surveys in operations research and management
  science\/}} \bibinfo{volume}{20}, \bibinfo{number}{1} (\bibinfo{year}{2015}),
  \bibinfo{pages}{1--18}.
\newblock


\bibitem[\protect\citeauthoryear{den Boer and Zwart}{den Boer and
  Zwart}{2013}]%
        {den2013simultaneously}
\bibfield{author}{\bibinfo{person}{Arnoud~V den Boer} {and}
  \bibinfo{person}{Bert Zwart}.} \bibinfo{year}{2013}\natexlab{}.
\newblock \showarticletitle{Simultaneously learning and optimizing using
  controlled variance pricing}.
\newblock \bibinfo{journal}{{\em Management science\/}} \bibinfo{volume}{60},
  \bibinfo{number}{3} (\bibinfo{year}{2013}), \bibinfo{pages}{770--783}.
\newblock


\bibitem[\protect\citeauthoryear{Farias and Van~Roy}{Farias and
  Van~Roy}{2010}]%
        {farias2010dynamic}
\bibfield{author}{\bibinfo{person}{Vivek~F Farias} {and}
  \bibinfo{person}{Benjamin Van~Roy}.} \bibinfo{year}{2010}\natexlab{}.
\newblock \showarticletitle{Dynamic pricing with a prior on market response}.
\newblock \bibinfo{journal}{{\em Operations Research\/}} \bibinfo{volume}{58},
  \bibinfo{number}{1} (\bibinfo{year}{2010}), \bibinfo{pages}{16--29}.
\newblock


\bibitem[\protect\citeauthoryear{Ferreira, Simchi-Levi, and Wang}{Ferreira
  et~al\mbox{.}}{2016}]%
        {ferreira2016online}
\bibfield{author}{\bibinfo{person}{Kris~Johnson Ferreira},
  \bibinfo{person}{David Simchi-Levi}, {and} \bibinfo{person}{He Wang}.}
  \bibinfo{year}{2016}\natexlab{}.
\newblock \showarticletitle{Online network revenue management using Thompson
  sampling}.
\newblock \bibinfo{journal}{{\em Harvard Business School Technology \&
  Operations Mgt. Unit Working Paper. Available at SSRN:
  https://ssrn.com/abstract=2588730\/}} (\bibinfo{year}{2016}).
\newblock


\bibitem[\protect\citeauthoryear{Gopalan, Mannor, and Mansour}{Gopalan
  et~al\mbox{.}}{2014}]%
        {gopalan2014thompson}
\bibfield{author}{\bibinfo{person}{Aditya Gopalan}, \bibinfo{person}{Shie
  Mannor}, {and} \bibinfo{person}{Yishay Mansour}.}
  \bibinfo{year}{2014}\natexlab{}.
\newblock \showarticletitle{Thompson sampling for complex online problems}. In
  \bibinfo{booktitle}{{\em International Conference on Machine Learning}}.
  \bibinfo{pages}{100--108}.
\newblock


\bibitem[\protect\citeauthoryear{Harrison, Keskin, and Zeevi}{Harrison
  et~al\mbox{.}}{2012}]%
        {harrison2012bayesian}
\bibfield{author}{\bibinfo{person}{J~Michael Harrison}, \bibinfo{person}{N~Bora
  Keskin}, {and} \bibinfo{person}{Assaf Zeevi}.}
  \bibinfo{year}{2012}\natexlab{}.
\newblock \showarticletitle{Bayesian dynamic pricing policies: Learning and
  earning under a binary prior distribution}.
\newblock \bibinfo{journal}{{\em Management Science\/}} \bibinfo{volume}{58},
  \bibinfo{number}{3} (\bibinfo{year}{2012}), \bibinfo{pages}{570--586}.
\newblock


\bibitem[\protect\citeauthoryear{Kawale, Bui, Kveton, Tran-Thanh, and
  Chawla}{Kawale et~al\mbox{.}}{2015}]%
        {kawale2015efficient}
\bibfield{author}{\bibinfo{person}{Jaya Kawale}, \bibinfo{person}{Hung~H Bui},
  \bibinfo{person}{Branislav Kveton}, \bibinfo{person}{Long Tran-Thanh}, {and}
  \bibinfo{person}{Sanjay Chawla}.} \bibinfo{year}{2015}\natexlab{}.
\newblock \showarticletitle{Efficient Thompson Sampling for Online￼
  Matrix-Factorization Recommendation}. In \bibinfo{booktitle}{{\em Advances in
  Neural Information Processing Systems}}. \bibinfo{pages}{1297--1305}.
\newblock


\bibitem[\protect\citeauthoryear{Keskin and Zeevi}{Keskin and Zeevi}{2014}]%
        {keskin2014dynamic}
\bibfield{author}{\bibinfo{person}{N~Bora Keskin} {and} \bibinfo{person}{Assaf
  Zeevi}.} \bibinfo{year}{2014}\natexlab{}.
\newblock \showarticletitle{Dynamic pricing with an unknown demand model:
  Asymptotically optimal semi-myopic policies}.
\newblock \bibinfo{journal}{{\em Operations Research\/}} \bibinfo{volume}{62},
  \bibinfo{number}{5} (\bibinfo{year}{2014}), \bibinfo{pages}{1142--1167}.
\newblock


\bibitem[\protect\citeauthoryear{Osband and Van~Roy}{Osband and
  Van~Roy}{2015}]%
        {osband2015bootstrapped}
\bibfield{author}{\bibinfo{person}{Ian Osband} {and} \bibinfo{person}{Benjamin
  Van~Roy}.} \bibinfo{year}{2015}\natexlab{}.
\newblock \showarticletitle{Bootstrapped thompson sampling and deep
  exploration}.
\newblock \bibinfo{journal}{{\em arXiv preprint arXiv:1507.00300\/}}
  (\bibinfo{year}{2015}).
\newblock


\bibitem[\protect\citeauthoryear{Phillips}{Phillips}{2005}]%
        {phillips2005pricing}
\bibfield{author}{\bibinfo{person}{Robert~Lewis Phillips}.}
  \bibinfo{year}{2005}\natexlab{}.
\newblock \bibinfo{booktitle}{{\em Pricing and revenue optimization}}.
\newblock \bibinfo{publisher}{Stanford University Press}.
\newblock


\bibitem[\protect\citeauthoryear{Rothschild}{Rothschild}{1974}]%
        {rothschild1974two}
\bibfield{author}{\bibinfo{person}{Michael Rothschild}.}
  \bibinfo{year}{1974}\natexlab{}.
\newblock \showarticletitle{A two-armed bandit theory of market pricing}.
\newblock \bibinfo{journal}{{\em Journal of Economic Theory\/}}
  \bibinfo{volume}{9}, \bibinfo{number}{2} (\bibinfo{year}{1974}),
  \bibinfo{pages}{185--202}.
\newblock


\bibitem[\protect\citeauthoryear{Russo, Van~Roy, Kazerouni, and Osband}{Russo
  et~al\mbox{.}}{2017}]%
        {russo2017tutorial}
\bibfield{author}{\bibinfo{person}{Daniel Russo}, \bibinfo{person}{Benjamin
  Van~Roy}, \bibinfo{person}{Abbas Kazerouni}, {and} \bibinfo{person}{Ian
  Osband}.} \bibinfo{year}{2017}\natexlab{}.
\newblock \showarticletitle{A Tutorial on Thompson Sampling}.
\newblock \bibinfo{journal}{{\em arXiv preprint arXiv:1707.02038\/}}
  (\bibinfo{year}{2017}).
\newblock


\bibitem[\protect\citeauthoryear{Shah, Yang, Alle, Ratnaparkhi, Shahshahani,
  and Chandra}{Shah et~al\mbox{.}}{2017}]%
        {shah2017practical}
\bibfield{author}{\bibinfo{person}{Parikshit Shah}, \bibinfo{person}{Ming
  Yang}, \bibinfo{person}{Sachidanand Alle}, \bibinfo{person}{Adwait
  Ratnaparkhi}, \bibinfo{person}{Ben Shahshahani}, {and} \bibinfo{person}{Rohit
  Chandra}.} \bibinfo{year}{2017}\natexlab{}.
\newblock \showarticletitle{A Practical Exploration System for Search
  Advertising}. In \bibinfo{booktitle}{{\em Proceedings of the 23rd ACM SIGKDD
  International Conference on Knowledge Discovery and Data Mining}}. ACM,
  \bibinfo{pages}{1625--1631}.
\newblock


\bibitem[\protect\citeauthoryear{Talluri and Van~Ryzin}{Talluri and
  Van~Ryzin}{2006}]%
        {talluri2006theory}
\bibfield{author}{\bibinfo{person}{Kalyan~T Talluri} {and}
  \bibinfo{person}{Garrett~J Van~Ryzin}.} \bibinfo{year}{2006}\natexlab{}.
\newblock \bibinfo{booktitle}{{\em The theory and practice of revenue
  management}}. Vol.~\bibinfo{volume}{68}.
\newblock \bibinfo{publisher}{Springer Science \& Business Media}.
\newblock


\bibitem[\protect\citeauthoryear{Thompson}{Thompson}{1933}]%
        {thompson1933likelihood}
\bibfield{author}{\bibinfo{person}{William~R Thompson}.}
  \bibinfo{year}{1933}\natexlab{}.
\newblock \showarticletitle{On the likelihood that one unknown probability
  exceeds another in view of the evidence of two samples}.
\newblock \bibinfo{journal}{{\em Biometrika\/}} \bibinfo{volume}{25},
  \bibinfo{number}{3/4} (\bibinfo{year}{1933}), \bibinfo{pages}{285--294}.
\newblock


\bibitem[\protect\citeauthoryear{Wasserman}{Wasserman}{2013}]%
        {wasserman2013all}
\bibfield{author}{\bibinfo{person}{Larry Wasserman}.}
  \bibinfo{year}{2013}\natexlab{}.
\newblock \bibinfo{booktitle}{{\em All of statistics: a concise course in
  statistical inference}}.
\newblock \bibinfo{publisher}{Springer Science \& Business Media}.
\newblock


\end{thebibliography}


\begin{thebibliography}{31}
\providecommand{\natexlab}[1]{#1}
\providecommand{\url}[1]{\texttt{#1}}
\expandafter\ifx\csname urlstyle\endcsname\relax
  \providecommand{\doi}[1]{doi: #1}\else
  \providecommand{\doi}{doi: \begingroup \urlstyle{rm}\Url}\fi

\bibitem[Agarwal et~al.(2016)Agarwal, Bird, Cozowicz, Hoang, Langford, Lee, Li,
  Melamed, Oshri, Ribas, et~al.]{agarwal2016multiworld}
Agarwal, Alekh, Bird, Sarah, Cozowicz, Markus, Hoang, Luong, Langford, John,
  Lee, Stephen, Li, Jiaji, Melamed, Dan, Oshri, Gal, Ribas, Oswaldo, et~al.
\newblock A multiworld testing decision service.
\newblock \emph{arXiv preprint arXiv:1606.03966}, 2016.

\bibitem[Aghion et~al.(1991)Aghion, Bolton, Harris, and
  Jullien]{aghion1991optimal}
Aghion, Philippe, Bolton, Patrick, Harris, Christopher, and Jullien, Bruno.
\newblock Optimal learning by experimentation.
\newblock \emph{The review of economic studies}, 58\penalty0 (4):\penalty0
  621--654, 1991.

\bibitem[Araman \& Caldentey(2009)Araman and Caldentey]{araman2009dynamic}
Araman, Victor~F and Caldentey, Ren{\'e}.
\newblock Dynamic pricing for nonperishable products with demand learning.
\newblock \emph{Operations Research}, 57\penalty0 (5):\penalty0 1169--1188,
  2009.

\bibitem[Auer et~al.(2002{\natexlab{a}})Auer, Cesa-Bianchi, and
  Fischer]{auer2002finite}
Auer, Peter, Cesa-Bianchi, Nicolo, and Fischer, Paul.
\newblock Finite-time analysis of the multiarmed bandit problem.
\newblock \emph{Machine learning}, 47\penalty0 (2-3):\penalty0 235--256,
  2002{\natexlab{a}}.

\bibitem[Auer et~al.(2002{\natexlab{b}})Auer, Cesa-Bianchi, Freund, and
  Schapire]{auer2002nonstochastic}
Auer, Peter, Cesa-Bianchi, Nicolo, Freund, Yoav, and Schapire, Robert~E.
\newblock The nonstochastic multiarmed bandit problem.
\newblock \emph{SIAM journal on computing}, 32\penalty0 (1):\penalty0 48--77,
  2002{\natexlab{b}}.

\bibitem[Aviv \& Pazgal(2005)Aviv and Pazgal]{aviv2005partially}
Aviv, Yossi and Pazgal, Amit.
\newblock A partially observed markov decision process for dynamic pricing.
\newblock \emph{Management Science}, 51\penalty0 (9):\penalty0 1400--1416,
  2005.

\bibitem[Ben-Tal et~al.(2009)Ben-Tal, El~Ghaoui, and Nemirovski]{ben2009robust}
Ben-Tal, Aharon, El~Ghaoui, Laurent, and Nemirovski, Arkadi.
\newblock \emph{Robust optimization}.
\newblock Princeton University Press, 2009.

\bibitem[Bertsekas \& Scientific(2015)Bertsekas and
  Scientific]{bertsekas2015convex}
Bertsekas, Dimitri~P and Scientific, Athena.
\newblock \emph{Convex optimization algorithms}.
\newblock Athena Scientific Belmont, 2015.

\bibitem[Besbes \& Zeevi(2009)Besbes and Zeevi]{besbes2009dynamic}
Besbes, Omar and Zeevi, Assaf.
\newblock Dynamic pricing without knowing the demand function: Risk bounds and
  near-optimal algorithms.
\newblock \emph{Operations Research}, 57\penalty0 (6):\penalty0 1407--1420,
  2009.

\bibitem[Blei et~al.(2017)Blei, Kucukelbir, and McAuliffe]{blei2017variational}
Blei, David~M, Kucukelbir, Alp, and McAuliffe, Jon~D.
\newblock Variational inference: A review for statisticians.
\newblock \emph{Journal of the American Statistical Association}, \penalty0
  (just-accepted), 2017.

\bibitem[Broder \& Rusmevichientong(2012)Broder and
  Rusmevichientong]{broder2012dynamic}
Broder, Josef and Rusmevichientong, Paat.
\newblock Dynamic pricing under a general parametric choice model.
\newblock \emph{Operations Research}, 60\penalty0 (4):\penalty0 965--980, 2012.

\bibitem[Bubeck et~al.(2012)Bubeck, Cesa-Bianchi, et~al.]{bubeck2012regret}
Bubeck, S{\'e}bastien, Cesa-Bianchi, Nicolo, et~al.
\newblock Regret analysis of stochastic and nonstochastic multi-armed bandit
  problems.
\newblock \emph{Foundations and Trends{\textregistered} in Machine Learning},
  5\penalty0 (1):\penalty0 1--122, 2012.

\bibitem[Caruana(1998)]{caruana1998multitask}
Caruana, Rich.
\newblock Multitask learning.
\newblock In \emph{Learning to learn}, pp.\  95--133. Springer, 1998.

\bibitem[Carvalho \& Puterman(2005{\natexlab{a}})Carvalho and
  Puterman]{carvalho2005learning}
Carvalho, Alexandre~X and Puterman, Martin~L.
\newblock Learning and pricing in an internet environment with binomial
  demands.
\newblock \emph{Journal of Revenue and Pricing Management}, 3\penalty0
  (4):\penalty0 320--336, 2005{\natexlab{a}}.

\bibitem[Carvalho \& Puterman(2005{\natexlab{b}})Carvalho and
  Puterman]{carvalho2015dynamic}
Carvalho, Alexandre~X and Puterman, Martin~L.
\newblock Dynamic optimization and learning: How should a manager set prices
  when the demand function is unknown?
\newblock 2005{\natexlab{b}}.

\bibitem[den Boer(2015)]{den2015dynamic}
den Boer, Arnoud~V.
\newblock Dynamic pricing and learning: historical origins, current research,
  and new directions.
\newblock \emph{Surveys in operations research and management science},
  20\penalty0 (1):\penalty0 1--18, 2015.

\bibitem[den Boer \& Zwart(2013)den Boer and Zwart]{den2013simultaneously}
den Boer, Arnoud~V and Zwart, Bert.
\newblock Simultaneously learning and optimizing using controlled variance
  pricing.
\newblock \emph{Management science}, 60\penalty0 (3):\penalty0 770--783, 2013.

\bibitem[Farias \& Van~Roy(2010)Farias and Van~Roy]{farias2010dynamic}
Farias, Vivek~F and Van~Roy, Benjamin.
\newblock Dynamic pricing with a prior on market response.
\newblock \emph{Operations Research}, 58\penalty0 (1):\penalty0 16--29, 2010.

\bibitem[Ferreira et~al.(2016)Ferreira, Simchi-Levi, and
  Wang]{ferreira2016online}
Ferreira, Kris~Johnson, Simchi-Levi, David, and Wang, He.
\newblock Online network revenue management using thompson sampling.
\newblock \emph{Harvard Business School Technology \& Operations Mgt. Unit
  Working Paper. Available at SSRN: https://ssrn.com/abstract=2588730}, 2016.

\bibitem[Gopalan et~al.(2014)Gopalan, Mannor, and Mansour]{gopalan2014thompson}
Gopalan, Aditya, Mannor, Shie, and Mansour, Yishay.
\newblock Thompson sampling for complex online problems.
\newblock In \emph{International Conference on Machine Learning}, pp.\
  100--108, 2014.

\bibitem[Harrison et~al.(2012)Harrison, Keskin, and
  Zeevi]{harrison2012bayesian}
Harrison, J~Michael, Keskin, N~Bora, and Zeevi, Assaf.
\newblock Bayesian dynamic pricing policies: Learning and earning under a
  binary prior distribution.
\newblock \emph{Management Science}, 58\penalty0 (3):\penalty0 570--586, 2012.

\bibitem[Kawale et~al.(2015)Kawale, Bui, Kveton, Tran-Thanh, and
  Chawla]{kawale2015efficient}
Kawale, Jaya, Bui, Hung~H, Kveton, Branislav, Tran-Thanh, Long, and Chawla,
  Sanjay.
\newblock Efficient thompson sampling for online￼ matrix-factorization
  recommendation.
\newblock In \emph{Advances in Neural Information Processing Systems}, pp.\
  1297--1305, 2015.

\bibitem[Keskin \& Zeevi(2014)Keskin and Zeevi]{keskin2014dynamic}
Keskin, N~Bora and Zeevi, Assaf.
\newblock Dynamic pricing with an unknown demand model: Asymptotically optimal
  semi-myopic policies.
\newblock \emph{Operations Research}, 62\penalty0 (5):\penalty0 1142--1167,
  2014.

\bibitem[Osband \& Van~Roy(2015)Osband and Van~Roy]{osband2015bootstrapped}
Osband, Ian and Van~Roy, Benjamin.
\newblock Bootstrapped thompson sampling and deep exploration.
\newblock \emph{arXiv preprint arXiv:1507.00300}, 2015.

\bibitem[Phillips(2005)]{phillips2005pricing}
Phillips, Robert~Lewis.
\newblock \emph{Pricing and revenue optimization}.
\newblock Stanford University Press, 2005.

\bibitem[Rothschild(1974)]{rothschild1974two}
Rothschild, Michael.
\newblock A two-armed bandit theory of market pricing.
\newblock \emph{Journal of Economic Theory}, 9\penalty0 (2):\penalty0 185--202,
  1974.

\bibitem[Russo et~al.(2017)Russo, Van~Roy, Kazerouni, and
  Osband]{russo2017tutorial}
Russo, Daniel, Van~Roy, Benjamin, Kazerouni, Abbas, and Osband, Ian.
\newblock A tutorial on thompson sampling.
\newblock \emph{arXiv preprint arXiv:1707.02038}, 2017.

\bibitem[Shah et~al.(2017)Shah, Yang, Alle, Ratnaparkhi, Shahshahani, and
  Chandra]{shah2017practical}
Shah, Parikshit, Yang, Ming, Alle, Sachidanand, Ratnaparkhi, Adwait,
  Shahshahani, Ben, and Chandra, Rohit.
\newblock A practical exploration system for search advertising.
\newblock In \emph{Proceedings of the 23rd ACM SIGKDD International Conference
  on Knowledge Discovery and Data Mining}, pp.\  1625--1631. ACM, 2017.

\bibitem[Talluri \& Van~Ryzin(2006)Talluri and Van~Ryzin]{talluri2006theory}
Talluri, Kalyan~T and Van~Ryzin, Garrett~J.
\newblock \emph{The theory and practice of revenue management}, volume~68.
\newblock Springer Science \& Business Media, 2006.

\bibitem[Thompson(1933)]{thompson1933likelihood}
Thompson, William~R.
\newblock On the likelihood that one unknown probability exceeds another in
  view of the evidence of two samples.
\newblock \emph{Biometrika}, 25\penalty0 (3/4):\penalty0 285--294, 1933.

\bibitem[Wasserman(2013)]{wasserman2013all}
Wasserman, Larry.
\newblock \emph{All of statistics: a concise course in statistical inference}.
\newblock Springer Science \& Business Media, 2013.

\end{thebibliography}

\end{document}